\documentclass[sigconf]{acmart}

\usepackage{CJKutf8}
\usepackage{amsmath}
\usepackage{algorithm}
\usepackage{enumitem}
\usepackage{picinpar}

\usepackage{graphicx}

\usepackage{algorithmicx}
\usepackage[noend]{algpseudocode}
\usepackage{subfigure}
\usepackage{multirow}
\usepackage{color}
\usepackage{balance}
\usepackage{enumitem}
\usepackage{hhline}
\usepackage[normalem]{ulem}
\usepackage{booktabs}
\usepackage{wrapfig}
\usepackage{cancel}
\usepackage{hyperref}
\usepackage{makecell}

\newcommand{\bL}{\ensuremath{\mathcal{L}}}

\newcommand{\bS}{\ensuremath{\mathcal{S}}}

\newcommand{\bG}{\ensuremath{\mathcal{G}}}
\newcommand{\bN}{\ensuremath{\mathcal{N}}}

\newcommand{\bT}{\ensuremath{\mathcal{T}}}

\newcommand{\bC}{\ensuremath{\mathcal{C}}}

\newcommand{\bD}{\ensuremath{\mathcal{D}}}

\renewcommand{\vec}[1]{\ensuremath{\mathbf{#1}}}

\newcommand{\stitle}[1]{\vspace{0.8mm} \noindent {\bf #1}}

\newcommand{\eg}{{\it e.g.}}

\newcommand{\ie}{{\it i.e.}}

\newcommand{\method}[1]{\textsc{#1}}
\newcommand{\model}{\method{GraphPrompt}{}}

\newcommand{\eat}[1]{}

\newcommand{\stkout}[1]{\ifmmode\text{\sout{\ensuremath{#1}}}\else\sout{#1}\fi}

\AtBeginDocument{%
  }

\setcopyright{acmcopyright}
\copyrightyear{2018}
\acmYear{2018}
\acmDOI{XXXXXXX.XXXXXXX}

\copyrightyear{2023} 
\acmYear{2023} 
\setcopyright{rightsretained} 
\acmConference[WWW '23]{Proceedings of the ACM Web Conference 2023}{May 1--5, 2023}{Austin, TX, USA}
\acmBooktitle{Proceedings of the ACM Web Conference 2023 (WWW '23), May 1--5, 2023, Austin, TX, USA}
\acmDOI{10.1145/3543507.3583386}
\acmISBN{978-1-4503-9416-1/23/04}

\begin{document}

\title[\model]{\model: Unifying Pre-Training and Downstream Tasks \\ for Graph Neural Networks}

\author{Zemin Liu$^{1*}$}
\affiliation{%
  \institution{$^{1}$National University of Singapore}
  \country{Singapore}
}
\email{zeminliu@nus.edu.sg}

\author{Xingtong Yu$^{2*}$}
\affiliation{%
 \institution{$^{2}$University of Science and Technology of China}
 \country{China}}
\email{yxt95@mail.ustc.edu.cn}

\author{Yuan Fang$^{3\dagger}$}
\affiliation{%
  \institution{$^{3}$Singapore Management University}
  \country{Singapore}}
\email{yfang@smu.edu.sg}

\author{Xinming Zhang$^{2\dagger}$}
\affiliation{%
  \institution{$^{2}$University of Science and Technology of China}
  \country{China}}
\email{xinming@ustc.edu.cn}

\thanks{
    $^*$Co-first authors with equal contribution. Part of the work was done while at Singapore Management University.\\
    $^{\dagger}$Corresponding authors.
}

\renewcommand{\shortauthors}{Zemin Liu, Xingtong Yu, Yuan Fang, Xinming Zhang}
 
\begin{abstract}
Graphs can model complex relationships between objects, enabling a myriad of Web applications such as online page/article classification and social recommendation. 
While graph neural networks (GNNs) have emerged as a powerful tool for graph representation learning, in an end-to-end supervised setting, their performance heavily relies on a large amount of task-specific supervision.
To reduce labeling requirement, the ``pre-train, fine-tune'' and ``pre-train, prompt'' paradigms have become increasingly common. In particular, prompting is a popular alternative to fine-tuning in natural language processing, which is designed to narrow the gap between pre-training and downstream objectives in a task-specific manner. However, existing study of prompting on graphs is still limited, lacking a universal 
treatment to appeal to different downstream tasks. In this paper, we propose \model, a novel pre-training and prompting framework on graphs. \model\ not only unifies pre-training and downstream tasks into a common task template, but also employs a learnable prompt to assist a downstream task in locating the most relevant knowledge from the pre-trained model in a task-specific manner. Finally, we conduct extensive experiments on five public datasets to evaluate and analyze \model.
\end{abstract}






\maketitle

\section{Introduction}

The ubiquitous Web is becoming the ultimate data repository, capable of linking a broad spectrum of objects to form gigantic and complex graphs. The prevalence of graph data enables a series of downstream tasks for Web applications, ranging from online page/article classification to friend recommendation in social networks.
Modern approaches for graph analysis generally resort to graph representation learning including graph embedding and graph neural networks (GNNs). Earlier graph embedding approaches \cite{perozzi2014deepwalk,tang2015line,grover2016node2vec} usually embed nodes on the graph into a low-dimensional space, in which the structural information such as the proximity between nodes can be captured \cite{cai2018comprehensive}.
More recently, GNNs \cite{kipf2016semi,hamilton2017inductive,velivckovic2017graph,xu2018powerful} have emerged as the state of the art for graph representation learning. Their key idea boils down to a message-passing framework, in which each node derives its representation by receiving and aggregating messages from its neighboring nodes recursively \cite{wu2020comprehensive}. 

\stitle{Graph pre-training.}
Typically, GNNs work in an end-to-end manner, and their performance depends heavily on the availability of large-scale, task-specific labeled data as supervision. This supervised paradigm presents two problems. First, task-specific supervision is often difficult or costly to obtain. Second, to deal with a new task, the weights of GNN models need to be retrained from scratch, even if the task is on the same graph.  
To address these issues, pre-training GNNs \cite{hu2020gpt,Hu2020Strategies,qiu2020gcc,lu2021learning} has become increasingly popular, inspired by pre-training techniques in language and vision applications \cite{dong2019unified,bao2022beit}. 
The pre-training of GNNs leverages self-supervised learning on more readily available label-free graphs (\ie, graphs without task-specific labels), and learns intrinsic graph properties that intend to be general across tasks and graphs in a domain. In other words, the pre-training extracts a task-agnostic prior, and can be used to initialize model weights for a new task. Subsequently, the initial weights can be quickly updated through a lightweight fine-tuning step on a smaller number of task-specific labels.


However, the ``pre-train, fine-tune'' paradigm suffers from the problem of inconsistent objectives between  pre-training and downstream tasks, resulting in suboptimal performance \cite{liu2021pre}. 
On one hand, the pre-training step aims to preserve various intrinsic graph properties such as node/edge features \cite{Hu2020Strategies,hu2020gpt}, node connectivity/links \cite{hamilton2017inductive,hu2020gpt,lu2021learning}, and local/global patterns \cite{qiu2020gcc,Hu2020Strategies,lu2021learning}.
On the other hand, the fine-tuning step aims to reduce the task loss, \ie, to fit the ground truth of the downstream task.
The discrepancy between the two steps can be quite large. 
For example, pre-training may focus on learning the connectivity pattern between two nodes (\ie, related to link prediction), whereas fine-tuning could be dealing with a node or graph property  (\ie, node classification or graph classification task). 


\stitle{Prior work.} To narrow the gap between pre-training and downstream tasks, prompting \cite{brown2020language} has first been proposed for language models, which is a natural language instruction designed for a specific downstream task to ``prompt out'' the semantic relevance between the task and the language model. Meanwhile, the parameters of the pre-trained language model are frozen without any fine-tuning, as the prompt can ``pull'' the task toward the pre-trained model. Thus, prompting is also more efficient than fine-tuning, especially when the pre-trained model is huge.  
Recently, prompting has also been introduced to graph pre-training in the GPPT approach  \cite{sun2022gppt}.
While the pioneering work has proposed a sophisticated design of pre-training and prompting, it can only be employed for the node classification task, lacking a universal treatment that  appeals to different downstream tasks  such as both node classification and graph classification.

%


\begin{figure}[t]
\centering
\includegraphics[width=0.99\linewidth]{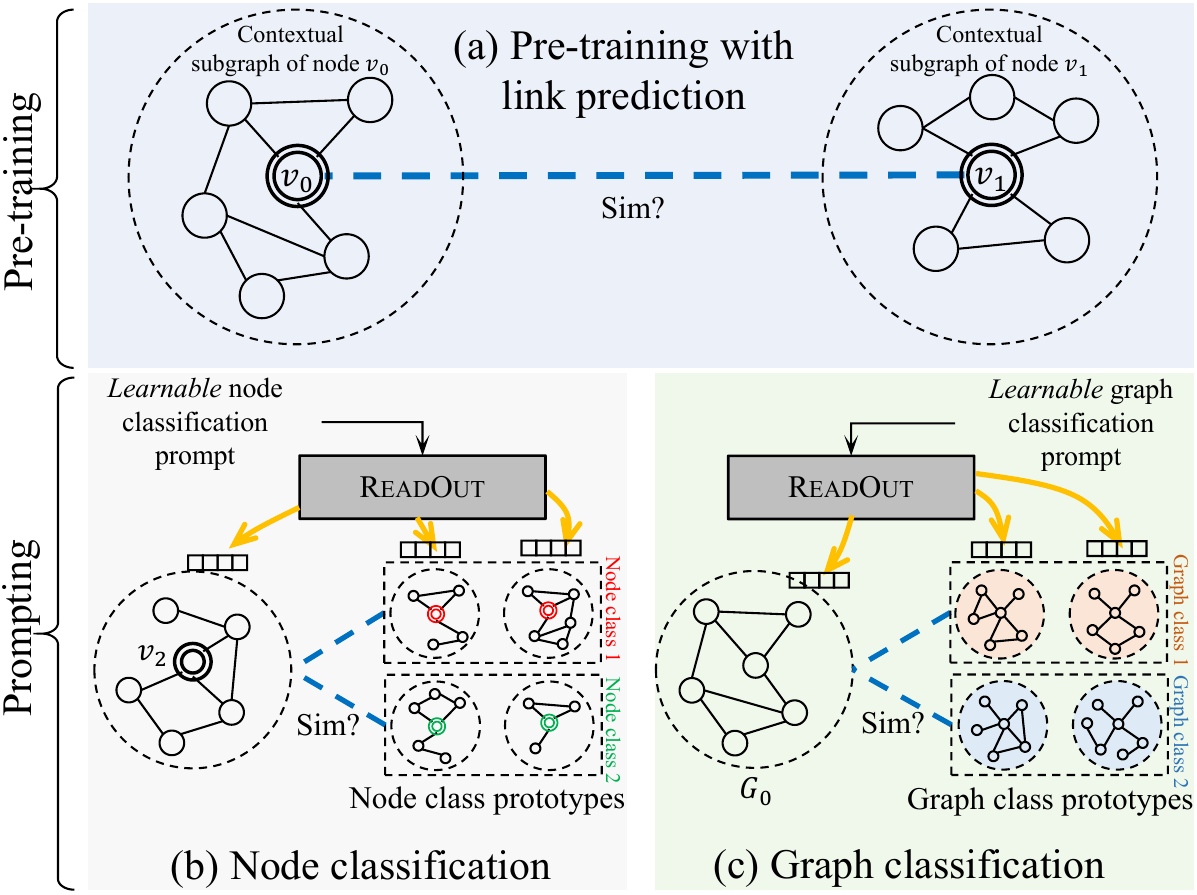}
\caption{Illustration of the motivation. (a) Pre-training on graphs. (b/c) Downstream node/graph classification.}
\label{fig.intro-motivation}
\end{figure}

\stitle{Research problem and challenges.}
To address the divergence between graph pre-training and various downstream tasks, in this paper we investigate the design of pre-training and prompting for graph neural networks. In particular, we aim for a unified design that can suit different downstream tasks flexibly.  
This problem is non-trivial due to the following two challenges.

Firstly, to enable effective knowledge transfer from the pre-training to a downstream task, it is desirable that the pre-training step preserves graph properties that are compatible with the given task. However, since different downstream tasks often have different objectives, \emph{how do we unify pre-training with various downstream tasks on graphs}, so that a single pre-trained model can universally support different tasks? That is, we try to convert the pre-training task and downstream tasks to follow the same ``template''. Using pre-trained language models as an analogy, both their pre-training and downstream tasks can be formulated as masked language modeling. 

Secondly, under the unification framework, it is still important to identify the distinction between different downstream tasks, in order to attain task-specific optima. For pre-trained language models, prompts in the form of natural language tokens or learnable word vectors have been designed to give different hints to different tasks, but it is less apparent what form prompts on graphs should take.  
Hence, \emph{how do we design prompts on graphs}, so that they can guide different downstream tasks to effectively make use of the pre-trained model?

\stitle{Present work.}
To address these challenges, we propose a novel graph pre-training and prompting framework, called  \model, aiming to unify the pre-training and downstream tasks for GNNs.
Drawing inspiration from the prompting strategy for pre-trained language models, \model\ capitalizes on a unified template to define the objectives for both pre-training and downstream tasks, thus bridging their gap. We further equip \model\ with task-specific learnable prompts, which guides the downstream task to exploit relevant knowledge from the pre-trained GNN model. 
The unified approach endows \model\ with the ability of working on limited supervision such as few-shot learning tasks.

More specifically, to address the first challenge of unification, we focus on graph topology, which is a key enabler of graph models. In particular, subgraph is a universal structure that can be leveraged for both node- and graph-level tasks. At the node level, the information of a node can be enriched and represented by its contextual subgraph, \ie, a subgraph where the node resides in \cite{zhang2018link,huang2020graph}; at the graph level, the information of a graph is naturally represented by the maximum subgraph (\ie, the graph itself).
Consequently, we unify both the node- and graph-level tasks, whether in pre-training or downstream, into the same template: the similarity calculation of (sub)graph\footnote{As a graph is a subgraph of itself, we may simply use subgraph to refer to a graph too.} representations. 
In this work, we adopt link prediction as the self-supervised pre-training task, given that links are readily available in any graph without additional annotation cost. Meanwhile, we focus on the popular node classification and graph classification as downstream tasks, which are node- and graph-level tasks, respectively. All these tasks can be cast as instances of learning subgraph similarity. On one hand, the link prediction task in pre-training boils down to the similarity  between the contextual subgraphs of two nodes, as shown in Fig.~\ref{fig.intro-motivation}(a). On the other hand, the downstream node or graph classification task boils down to the similarity between the target instance (a node's contextual subgraph or the whole graph, resp.) and the class prototypical subgraphs constructed from labeled data, as illustrated in Figs.~\ref{fig.intro-motivation}(b) and (c). The unified template bridges the gap between the pre-training and different downstream tasks. 


Toward the second challenge, we distinguish different downstream tasks by way of the $\textsc{ReadOut}$ operation on subgraphs. The $\textsc{ReadOut}$ operation is essentially an aggregation function to fuse node representations in the subgraph into a single subgraph representation. For instance, sum pooling, which sums the representations of all nodes in the subgraph, is a practical and popular scheme for $\textsc{ReadOut}$. However, different downstream tasks can benefit from different aggregation schemes for their $\textsc{ReadOut}$. In particular, node classification tends to focus on features that can contribute to the representation of the target node, while  graph classification tends to focus on features associated with the graph class.
%
Motivated by such differences, we propose a novel task-specific learnable prompt to guide the $\textsc{ReadOut}$ operation of each downstream task with an appropriate aggregation scheme.
As shown in Fig.~\ref{fig.intro-motivation}, the learnable prompt serves as the parameters of the $\textsc{ReadOut}$ operation of downstream tasks, and thus enables different aggregation functions on the subgraphs of different tasks. 
Hence, \model\ not only unifies the pre-training and downstream tasks into the same template based on subgraph similarity, but also recognizes the differences between various downstream tasks to guide task-specific objectives.

\stitle{Contributions.}
To summarize, our contributions are three-fold.
(1) We recognize the gap between graph pre-training and downstream tasks, and propose a unification framework \model\ based on subgraph similarity for both pre-training and downstream tasks, including both node and graph classification tasks.
(2) We propose a novel prompting strategy for \model, hinging on a learnable prompt to actively guide downstream tasks using task-specific aggregation in $\textsc{ReadOut}$, in order to drive the downstream tasks to exploit the pre-trained model in a task-specific manner.
(3) We conduct extensive experiments on five public datasets, and the results demonstrate the superior performance of \model\ in comparison to the state-of-the-art approaches. 


\section{Related Work}


\stitle{Graph representation learning.}
The rise of graph representation learning, including earlier graph embedding \cite{perozzi2014deepwalk,tang2015line,grover2016node2vec} and recent GNNs \cite{kipf2016semi,hamilton2017inductive,velivckovic2017graph,xu2018powerful}, opens up great opportunities for various downstream tasks at node and graph levels.
Note that learning graph-level representations requires an additional $\textsc{ReadOut}$ operation 
which summarizes the global information of a graph by aggregating node representations through a flat 
\cite{duvenaud2015convolutional,gilmer2017neural,zhang2018end,xu2018powerful} or hierarchical \cite{gao2019graph,lee2019self,ma2019graph,ying2018hierarchical} pooling algorithm.
We refer the readers to two comprehensive surveys \cite{cai2018comprehensive,wu2020comprehensive} for more details.

\stitle{Graph pre-training.}
Inspired by the application of pre-training models in language \cite{dong2019unified,beltagy2019scibert} and vision \cite{lu2019vilbert,bao2022beit} domains, graph pre-training \cite{xia2022survey} emerges as a powerful paradigm that leverages self-supervision on label-free graphs to learn intrinsic graph properties. 
While the pre-training learns a task-agnostic prior, a relatively light-weight fine-tuning step is further employed to update the pre-trained weights to fit a given downstream task. 
Different pre-training approaches 
design different self-supervised tasks based on various graph properties such as node features \cite{Hu2020Strategies,hu2020gpt}, links \cite{kipf2016variational,hamilton2017inductive,hu2020gpt,lu2021learning}, local or global patterns \cite{qiu2020gcc,Hu2020Strategies,lu2021learning}, local-global consistency \cite{velickovic2019deep,Sun2020InfoGraph,peng2020graph,hassani2020contrastive},
and their combinations \cite{you2020graph,you2021graph,suresh2021adversarial}.

However, the above approaches do not consider the gap between pre-training and downstream objectives, which limits their generalization ability to handle different tasks.
Some recent studies recognize the importance of narrowing this gap. L2P-GNN \cite{lu2021learning} capitalizes on meta-learning \cite{finn2017model} to simulate the fine-tuning step during pre-training. However, since the downstream tasks can still differ from the simulation task, the problem is not fundamentally addressed.
In other fields, as an alternative to fine-tuning, researchers turn to prompting \cite{brown2020language}, in which a task-specific prompt is used to cue the downstream tasks. Prompts can be either handcrafted \cite{brown2020language} or learnable  \cite{liu2021gpt,lester2021power}. On graph data, the study of prompting is still limited. One recent work called GPPT \cite{sun2022gppt} capitalizes on a sophisticated design of learnable prompts on graphs, but it only works with node classification, lacking a unification effort to accommodate other downstream tasks like graph classification.
Besides, there is a model also named as GraphPrompt \cite{zhang2021graphprompt}, but it considers an NLP task (biomedical entity normalization) on text data, where graph is only auxiliary. It employs the standard text prompt unified by masked language modeling, assisted by a relational graph to generate text templates, which is distinct from our work.

\stitle{Comparison to other settings.}
Our few-shot setting is different from other paradigms that also deal with label scarcity, including semi-supervised learning \cite{kipf2016semi} and meta-learning \cite{finn2017model}. In particular, semi-supervised learning cannot cope with novel classes not seen in training, while meta-learning requires a large volume of labeled data in their base classes for a meta-training phase, before they can handle few-shot tasks in testing.  






\section{Preliminaries}

In this section, we give the problem definition and introduce the background of GNNs.

\subsection{Problem Definition}\label{sec:prelim:problem}

\stitle{Graph.}
A graph can be defined as $G=(V,E)$, where $V$ is the set of nodes and $E$ is the set of edges. We also assume an input feature matrix of the nodes, $\vec{X}\in\mathbb{R}^{|V|\times d}$, is available. Let $\vec{x}_i\in\mathbb{R}^d$ denote the feature vector of node $v_i\in V$. 
In addition, we denote a set of graphs as $\bG=\{G_1, G_2, \ldots, G_N\}$.

\stitle{Problem.}
In this paper, we investigate the problem of graph pre-training and prompting. For the downstream tasks, we consider the popular node classification and graph classification tasks.
For node classification on a graph $G=(V,E)$, let $C$ be the set of node classes with $\ell_i\in C$ denoting the class label of node $v_i \in V$.
For graph classification on a set of graphs $\bG$, let $\bC$ be the set of graph labels with $L_i\in \bC$ denoting the class label of graph $G_i\in \bG$.

In particular, the downstream tasks are given limited supervision in a few-shot setting: for each class in the two tasks, only $k$ labeled samples (\ie, nodes or graphs) are provided, known as $k$-shot classification.

\subsection{Graph Neural Networks}\label{sec:prelim:gnn}
The success of GNNs boils down to the message-passing mechanism \cite{wu2020comprehensive}, 
in which each node receives and aggregates messages (\ie, features or embeddings) from its neighboring nodes to generate its own representation. This operation of neighborhood aggregation can be stacked in multiple layers to enable recursive message passing. Formally, in the $l$-th GNN layer, the embedding of node $v$, denoted by $\vec{h}^l_v$, is calculated based on the embeddings in the previous layer, as follows.
\begin{equation}
    \vec{h}^l_v = \textsc{Aggr}(\vec{h}^{l-1}_v, \{\vec{h}^{l-1}_u : u\in\bN_v\}; \theta^l),
\end{equation}
where $\bN_v$ is the set of neighboring nodes of $v$, $\theta^l$ is the learnable GNN parameters in layer $l$. $\textsc{Aggr}(\cdot)$ is the neighborhood aggregation function and can take various forms, ranging from the simple mean pooling \cite{kipf2016semi,hamilton2017inductive} to advanced neural networks such as neural attention \cite{velivckovic2017graph} or multi-layer perceptrons \cite{xu2018powerful}. Note that in the first layer, the input node embedding $\vec{h}^0_v$ can be initialized as the node features in $\vec{X}$. 
The total learnable GNN parameters can be denoted as $\Theta=\{\theta^1, \theta^2, \ldots\}$. For brevity, we simply denote the output node representations of the last layer as $\vec{h}_v$.

\section{Proposed Approach}

In this section, we present our proposed approach \model.

\begin{figure*}[t]
\centering
\includegraphics[width=1\linewidth]{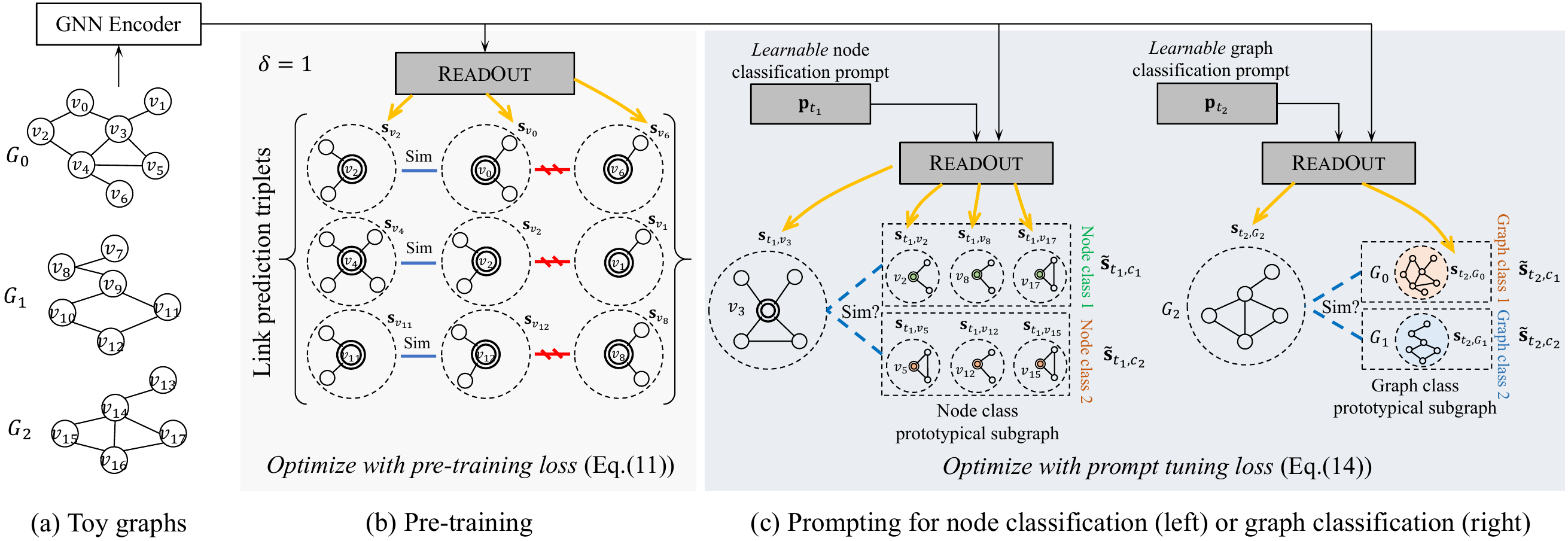}
\vspace{-6mm}
\caption{Overall framework of \model.}
\label{fig.framework}
\end{figure*}

\subsection{Unification Framework}\label{sec:model:unification}


We first introduce the overall  framework of \model\ in Fig.~\ref{fig.framework}. 
Our framework is deployed on a set of label-free graphs shown in  Fig.~\ref{fig.framework}(a), for pre-training in  Fig.~\ref{fig.framework}(b). 
The pre-training adopts a link prediction task, which is self-supervised without requiring extra annotation. 
Afterward, in Fig.~\ref{fig.framework}(c), we capitalize on a learnable prompt to guide each downstream task, namely, node classification or graph classification, for task-specific exploitation of the pre-trained model.
In the following, we explain how the framework supports a unified view of pre-training and downstream tasks.

\stitle{Instances as subgraphs.} The key to the unification of pre-training and downstream tasks lies in finding a common template for the tasks. The task-specific prompt can then be further fused with the template of each downstream task, to distinguish the varying characteristics of different tasks.



In comparison to other fields such as visual and language processing, 
graph learning is uniquely characterized by the exploitation of graph topology.
In particular, subgraph is a universal structure capable of expressing both node- and graph-level instances. 
On one hand, at the node level, every node resides in a local neighborhood, which in turn contextualizes the node \cite{DBLP:conf/cikm/LiuZFZH20,liu2021nodewise,liu2021tail}. The local neighborhood of a node $v$ on a graph $G=(V,E)$ is usually defined by a contextual subgraph $S_v=(V(S_v), E(S_v))$, where its set of nodes and edges are respectively given by
\begin{align}
V(S_v)&=\{d(u,v)\le \delta \mid u\in V\}, \text{ and }\\
E(S_v)&=\{ (u,u')\in E \mid u \in V(S_v), u'\in V(S_v) \},
\end{align}
where $d(u,v)$ gives the shortest distance between nodes $u$ and $v$ on the graph $G$, and $\delta$ is a predetermined threshold. That is, $S_v$ consists of nodes within $\delta$ hops from the node $v$, and the edges between those nodes. Thus, the contextual subgraph $S_v$ embodies not only the self-information of the node $v$, but also rich contextual information to complement the self-information \cite{zhang2018link,huang2020graph}.
On the other hand, at the graph level, the maximum subgraph of a graph $G$, denoted $S_{G}$, is the graph itself, \ie, $S_{G}=G$. The maximum subgraph  $S_{G}$ spontaneously embodies all information of $G$.
In summary, subgraphs can be used to represent both node- and graph-level instances: Given an instance $x$ which can either be a node or a graph (\eg, $x=v$ or $x=G$), the subgraph $S_x$ offers a unified access to the information associated with $x$.



\stitle{Unified task template.} Based on the above subgraph definitions for both node- and graph-level instances, we are ready to unify different tasks to follow a common template. Specifically, the link prediction task in pre-training and the downstream node and graph classification tasks can all be redefined as \emph{subgraph similarity learning}. Let $\vec{s}_x$ be the vector representation of the subgraph $S_x$, and $\text{sim}(\cdot,\cdot)$ be the cosine similarity function. As illustrated in Figs.~\ref{fig.framework}(b) and (c), the three tasks can be mapped to the computation  of subgraph similarity, which is formalized below. 
\begin{itemize}[leftmargin=*]
    \item \textbf{\textit{Link prediction}}: This is a node-level task. Given a graph $G=(V,E)$ and a triplet of nodes $(v,a,b)$ such that $(v,a)\in E$ and $(v,b)\notin E$, we shall have
    \begin{align}
    \text{sim}(\vec{s}_v,\vec{s}_a)>\text{sim}(\vec{s}_v,\vec{s}_b).
    \end{align}
    Intuitively, the contextual subgraph of $v$ shall be more similar to that of a node linked to $v$ than that of another unlinked node.
    
    \item \textbf{\textit{Node classification}}: This is also a node-level task. Consider a graph $G=(V,E)$ with a set of  node classes $C$, and a set of labeled nodes $D=\{(v_1,\ell_1),(v_2,\ell_2),\ldots\}$ where $v_i \in V$ and  $\ell_i$ is the corresponding label of $v_i$. As we adopt a $k$-shot setting, there are exactly $k$ pairs of $(v_i,\ell_i=c) \in D$ for every class $c\in C$.  
    For each class $c \in C$, further define a \emph{node class prototypical subgraph} represented by a vector $\vec{\tilde{s}}_c$, given by  
    \begin{align}
        \vec{\tilde{s}}_c = \frac{1}{k}\sum_{(v_i,\ell_i)\in D, \ell_i=c} \vec{s}_{v_i}.
    \end{align}
    Note that the class prototypical subgraph is a ``virtual'' subgraph with a latent representation in the same embedding space as the node contextual subgraphs. Basically, it is constructed as the mean representation of the contextual subgraphs of labeled nodes in a given class. 
    Then, given a node $v_j$ not in the labeled set $D$, its class label $\ell_j$ shall be 
    \begin{align}
        \ell_j = \arg\max_{c\in C} \text{sim}(\vec{s}_{v_j}, \vec{\tilde{s}}_c).
    \end{align}
    Intuitively, a node shall belong to the class whose prototypical subgraph is the most similar to the node's contextual subgraph.
    
    \item \textbf{\textit{Graph classification}}: This is a graph-level task. Consider a set of graphs $\bG$ with a set of graph classes $\bC$, and a set of labeled graphs $\bD=\{(G_1,L_1),(G_2,L_2),\ldots\}$ where $G_i \in \bG$ and  $L_i$ is the corresponding label of $G_i$. In the $k$-shot setting, there are exactly $k$ pairs of $(G_i,L_i=c) \in \bD$ for every class $c\in \bC$.  
    Similar to node classification, for each class $c \in \bC$, we define a \emph{graph class prototypical subgraph}, also represented by the mean embedding vector of the (sub)graphs in $c$:
    \begin{align}
        \vec{\tilde{s}}_c = \frac{1}{k}\sum_{(G_i,L_i)\in \bD, L_i=c} \vec{s}_{G_i}.
    \end{align}
    Then, given a graph $G_j$ not in the labeled set $\bD$, its class label $L_j$ shall be 
    \begin{align}
        L_j = \arg\max_{c\in \bC} \text{sim}(\vec{s}_{G_j}, \vec{\tilde{s}}_c).
    \end{align}
    Intuitively, a graph shall belong to the class whose prototypical subgraph is the most similar to itself. \hspace*{\fill}\qed
\end{itemize}

It is worth noting that node and graph classification can be further condensed into a single set of notations. Let $(x,y)$ be an annotated instance of graph data, \ie, $x$ is either a node or a graph, and $y\in Y$ is the class label of $x$ among the set of classes $Y$. Then,  
\begin{align}
        y = \arg\max_{c\in Y} \text{sim}(\vec{s}_{x}, \vec{\tilde{s}}_c).
    \end{align}



Finally, to materialize the common task template, we discuss how to learn the subgraph embedding vector $\vec{s}_x$ for the subgraph $S_x$. 
Given node representations $\vec{h}_{v}$ generated by a GNN (see Sect.~\ref{sec:prelim:gnn}), a standard approach of computing $\vec{s}_x$ is to employ a $\textsc{ReadOut}$ operation that aggregates the representations of nodes in the subgraph $S_x$. That is,
\begin{align} \label{eq.readout}
    \vec{s}_{x} = \textsc{ReadOut}(\{\vec{h}_v:v\in V(S_x)\}).
\end{align}
The choice of the aggregation scheme for $\textsc{ReadOut}$ is flexible, including sum pooling and more advanced techniques \cite{ying2018hierarchical,xu2018powerful}. In our implementation, we simply use sum pooling.

In summary, the unification framework is enabled by the common task template of subgraph similarity learning, which lays the foundation of our pre-training and prompting strategies as we will introduce in the following parts.



\subsection{Pre-Training Phase}

As discussed earlier, our pre-training phase employs the link prediction task. Using link prediction/generation is a popular and natural way \cite{hamilton2017inductive,hu2020gpt,DBLP:conf/nips/HwangPKKHK20,lu2021learning},
as a vast number of links are readily available on large-scale graph data without extra annotation. In other words, the link prediction objective can be optimized on label-free graphs, such as those shown in Fig.~\ref{fig.framework}(a), in a self-supervised manner.  

Based on the common template defined in Sect.~\ref{sec:model:unification}, the link prediction task is anchored on the similarity of the contextual subgraphs of two candidate nodes. Generally, the subgraphs of two positive (\ie, linked) candidates shall be more similar than those of negative (\ie, non-linked) candidates, as illustrated in Fig.~\ref{fig.framework}(b).
Subsequently, the pre-trained prior on subgraph similarity can be naturally transferred to node classification downstream, which shares a similar intuition: the subgraphs of nodes in the same class shall be more similar than those of nodes from different classes. On the other hand, the prior can also support graph classification downstream, as graph similarity is consistent with subgraph similarity not only in letter (as a graph is technically always a subgraph of itself), but also in spirit. The ``spirit'' here refers to the tendency that graphs sharing similar subgraphs are likely to be similar themselves, which means graph similarity can be translated into the similarity of the containing subgraphs \cite{shervashidze2011weisfeiler,zhang2018end,togninalli2019wasserstein}.   

Formally, given a node $v$ on graph $G$, we randomly sample one positive node $a$ from $v$'s neighbors, and a negative node $b$ from the graph that does not link to $v$, forming a triplet $(v,a,b)$. Our objective is to increase the similarity between the contextual subgraphs $S_v$ and $S_a$, while decreasing that between $S_v$ and $S_b$. More generally, on a set of label-free graphs $\bG$, we sample a number of triplets from each graph to construct an overall training set $\bT_\text{pre}$. Then, we define the following pre-training loss.
\begin{align} \label{eq.pre-train-loss}
    \bL_{\text{pre}}(\Theta)=-\sum_{(v,a,b)\in\bT_\text{pre}}\ln\frac{\exp(\text{sim}(\vec{s}_v,\vec{s}_a)/\tau)}{\sum_{u\in\{a,b\}}\exp(\text{sim}(\vec{s}_v,\vec{s}_u)/\tau)},
\end{align}
where 
$\tau$ is a temperature hyperparameter to control the shape of the output distribution. Note that the loss is parameterized by $\Theta$, which represents the GNN model weights. 

The output of the pre-training phase is the optimal model parameters $\Theta_0=\arg\min_\Theta \bL_{\text{pre}}(\Theta)$. $\Theta_0$ can be used to initialize the GNN weights for downstream tasks, thus enabling the transfer of prior knowledge downstream.





\subsection{Prompting for Downstream Tasks}

The unification of pre-training and downstream tasks enables more effective knowledge transfer as the tasks in the two phases are made more compatible by following a common template. However, it is still important to distinguish different downstream tasks, in order to capture task individuality and achieve task-specific optimum.

To cope with this challenge, we propose a novel task-specific learnable prompt on graphs, inspired by prompting in natural language processing \cite{brown2020language}. In language contexts, a prompt is initially a handcrafted instruction to guide the downstream task, which provides task-specific cues to extract relevant prior knowledge through a unified task template (typically, pre-training and downstream tasks are all mapped to masked language modeling). More recently, learnable prompts \cite{liu2021gpt,lester2021power} have been proposed as an alternative to handcrafted prompts, to alleviate the high engineering cost of the latter. 

\stitle{Prompt design.} Nevertheless, our proposal is distinctive from language-based prompting for two reasons.
Firstly, we have a different task template from masked language modeling. Secondly, since our prompts are designed for graph structures, they are more abstract and cannot take the form of language-based instructions. Thus, they are virtually impossible to be handcrafted. Instead, they should be topology related to align with the core of graph learning. In particular, under the same task template of subgraph similarity learning, 
the $\textsc{ReadOut}$ operation (used to generate the subgraph representation) can be ``prompted'' differently for different downstream tasks. Intuitively, different tasks can benefit
from different aggregation schemes for their $\textsc{ReadOut}$. For instance, node classification pays more attention to features that are topically more relevant to  the target node. In contrast, graph classification tends to focus on features that are correlated to the graph class. Moreover, the important features may also vary given different sets of instances or classes in a task.

Formally, let $\vec{p}_t$ denote a learnable \emph{prompt vector} for a downstream task $t$, as shown in Fig.~\ref{fig.framework}(c). The prompt-assisted $\textsc{ReadOut}$ operation on a subgraph $S_x$ for task $t$ is
\begin{align}\label{eq:prompt-fw}
\vec{s}_{t,x} = \textsc{ReadOut}(\{\vec{p}_t\odot\vec{h}_v:v\in V(S_x)\}),
\end{align}
where $\vec{s}_{t,x}$ is the task $t$-specific subgraph representation, and $\odot$ denotes the element-wise multiplication. That is, we perform a \emph{feature weighted} summation of the node representations from the subgraph, where the prompt vector $\vec{p}_t$ is a dimension-wise reweighting in order to extract the most relevant prior knowledge for the task $t$.

Note that other prompt designs are also possible. For example, we could consider a learnable \emph{prompt matrix} $\vec{P}_t$, which applies a linear transformation to the node representations:
\begin{align}\label{eq:prompt-lin}
\vec{s}_{t,x} = \textsc{ReadOut}(\{\vec{P}_t\vec{h}_v:v\in V(S_x)\}).
\end{align}
More complex prompts such as an attention layer is another alternative.
However, one of the main motivation of prompting instead of fine-tuning is to reduce reliance on labeled data. In few-shot settings, given very limited supervision, prompts with fewer parameters are preferred to mitigate the risk of overfitting. Hence, the feature weighting scheme in Eq.~\eqref{eq:prompt-fw} is adopted for our prompting as the prompt is a single vector of the same length as the node representation, which is typically a small number (\eg, 128). 



\stitle{Prompt tuning.}
To optimize the learnable prompt, also known as \emph{prompt tuning}, we formulate the loss based on the common template of subgraph similarity, using the prompt-assisted task-specific subgraph representations. 

Formally, consider a task $t$ with a labeled training set $\bT_{t} =\{(x_1,y_1),(x_2,y_2),\ldots\}$, where $x_i$ is an instance (\ie, a node or a graph), and $y_i\in Y$ is the class label of $x_i$ among the set of classes $Y$. The loss for prompt tuning is defined as
\begin{align}
    \bL_{\text{prompt}}(\vec{p}_t)=-\sum_{(x_i,y_i)\in \bT_t}\ln\frac{\exp(\text{sim}(\vec{s}_{t,x_i},\tilde{\vec{s}}_{t,y_i})/\tau)}{\sum_{c\in Y}\exp(\text{sim}(\vec{s}_{t,x_i},\tilde{\vec{s}}_{t,c})/\tau)},
\end{align}
where the class prototypical subgraph for class $c$ is represented by $\tilde{\vec{s}}_{t,c}$, which is also generated by the prompt-assisted, task-specific $\textsc{ReadOut}$.

Note that, the prompt tuning loss is only parameterized by the learnable prompt vector $\vec{p}_t$, without the GNN weights. Instead, the pre-trained GNN weights   $\Theta_0$ are frozen for downstream tasks, as no fine-tuning is necessary. This significantly decreases the number of parameters to be updated downstream, thus not only improving the computational efficiency of task learning and inference, but also reducing the reliance on labeled data.

\section{Experiments}
In this section, we conduct extensive experiments including node classification and graph classification as downstream tasks on five benchmark datasets to evaluate the proposed \model.

\subsection{Experimental Setup}

\stitle{Datasets.}
We employ five benchmark datasets for evaluation.
(1) \emph{Flickr} \cite{wen2021meta} is an image sharing network.
(2) \emph{PROTEINS} \cite{borgwardt2005protein} is a collection of protein graphs which include the amino acid sequence, conformation, structure, and features such as active sites of the proteins. 
(3) \emph{COX2} \cite{nr} is a dataset of molecular structures including 467 cyclooxygenase-2 inhibitors.
(4) \emph{ENZYMES} \cite{wang2022faith} is a dataset of 600 enzymes collected from the BRENDA enzyme database. 
(5) \emph{BZR} \cite{nr} is a collection of 405 ligands for benzodiazepine receptor. 

We summarize these datasets in Table~\ref{table.datasets}, and present further details in Appendix~\ref{app.dataset}.
Note that the ``Task'' column indicates the type of downstream task performed on each dataset: ``N'' for node classification and ``G'' for graph classification.


\begin{table}[tbp]
\center
\addtolength{\tabcolsep}{-0.5mm}
\caption{Summary of datasets. 
\label{table.datasets}}
\vspace{-2mm}
\small
\begin{tabular}{@{}c|rrrrrrc@{}}
\toprule
	& \makecell[c]{Graphs} & \makecell[c]{Graph \\ classes} & \makecell[c]{Avg.\\ nodes} & \makecell[c]{Avg. \\ edges} &  \makecell[c]{Node \\ features} &  \makecell[c]{Node \\ classes} & \makecell[c]{Task \\ (N/G)} \\
\midrule
     Flickr & 1 & - & 89,250 & 899,756 & 500 & 7 & N \\ 
     PROTEINS & 1,113 & 2 & 39.06 & 72.82 & 1 & 3 & N, G\\
     COX2 & 467 & 2 & 41.22 & 43.45 & 3 & - & G\\
     ENZYMES & 600 & 6 & 32.63 & 62.14 & 18 & 3 & N, G\\
     BZR & 405 & 2 & 35.75 & 38.36 & 3 & - & G\\
 \bottomrule
\end{tabular}
\end{table}

\stitle{Baselines.}
We evaluate \model\ against the state-of-the-art approaches from three main categories, as follows.
(1) \emph{End-to-end graph neural networks}: GCN \cite{kipf2016semi}, GraphSAGE \cite{hamilton2017inductive}, GAT \cite{velivckovic2017graph} and GIN \cite{xu2018powerful}. They capitalize on the key operation of neighborhood aggregation to recursively aggregate messages from the neighbors, and work in an end-to-end manner.
(2) \emph{Graph pre-training models}: DGI \cite{velickovic2019deep}, InfoGraph \cite{sun2019infograph}, and GraphCL \cite{you2020graph}. They work in the ``pre-train, fine-tune'' paradigm. In particular, they pre-train the GNN models to preserve the intrinsic graph properties, and fine-tune the pre-trained weights on downstream tasks to fit task labels.
(3) \emph{Graph prompt models}: GPPT \cite{sun2022gppt}. GPPT utilizes a link prediction task for pre-training, and resorts to a learnable prompt for the node classification task, which is mapped to a link prediction task.

Note that other few-shot learning methods on graphs, such as Meta-GNN \cite{zhou2019meta} and RALE \cite{liu2021relative}, adopt a meta-learning paradigm \cite{finn2017model}. Thus, they cannot be used in our setting, as they require labeled data in their base classes for the meta-training phase. In our approach, only label-free graphs are utilized for pre-training.

\stitle{Settings and parameters.}
To evaluate the goal of our \model\ in realizing a unified design that can suit different downstream tasks flexibly, we consider two typical types of downstream tasks, \ie, node classification and graph classification. 
In particular, for the datasets which are suitable for both of these two tasks, \ie, \emph{PROTEINS} and \emph{ENZYMES}, we only pre-train the GNN model once on each dataset, and utilize the same pre-trained model for the two downstream tasks with their task-specific prompting.

The downstream tasks follow a $k$-shot classification setting. For each type of downstream task, we construct a series of $k$-shot classification tasks. The details of task construction will be elaborated later when reporting the results in Sect.~\ref{sec:expt:perf}.
For task evaluation, as the $k$-shot tasks are balanced classification, we employ accuracy as the evaluation metric following earlier work \cite{wang2020graph,liu2021relative}.

For all the baselines, based on the authors' code and default settings, we further tune their hyper-parameters to optimize their performance. We present more implementation details of the baselines and our \model\ in Appendix~\ref{app.setting-parameter}.

\subsection{Performance Evaluation}\label{sec:expt:perf}

As discussed, we perform two types of downstream task different from the link prediction task in pre-training, namely, node classification and graph classification in few-shot settings. We first evaluate on a fixed-shot setting, and then vary the shot numbers to see the performance trend.

\stitle{Few-shot node classification.}
We conduct this node-level task on three datasets, \ie, \emph{Flickr}, \emph{PROTEINS}, and \emph{ENZYMES}. Following a typical $k$-shot setup \cite{zhou2019meta,wang2020graph,liu2021relative}, we generate a series of few-shot tasks for model training and validation. 
In particular, for \emph{PROTEINS} and \emph{ENZYMES}, on each graph we randomly generate ten 1-shot node classification tasks (\ie, in each task, we randomly sample 1 node per class) for training and validation, respectively. Each training task is paired with a validation task, and the remaining nodes not sampled by the pair of training and validation tasks will be used for testing. For \emph{Flickr}, as it contains a large number of very sparse node features, selecting very few shots for training may result in inferior performance for all the methods.
Therefore, we randomly generate ten 50-shot node classifcation tasks, for training and validation, respectively. On Flickr, 50 shots are still considered few, accounting for less than 0.06\% of all nodes on the graph.  

Table~\ref{table.node-classification} illustrates the results of few-shot node classification. We have the following observations. 
First, our proposed \model\ outperforms all the baselines across the three datasets, demonstrating the effectiveness of \model\ in transferring knowledge from the pre-training to downstream tasks.
In particular, by virtue of the unification framework and prompt-based task-specific aggregation in $\textsc{ReadOut}$ function, \model\ is able to narrow the gap between pre-training and downstream tasks, and guide the downstream tasks to exploit the pre-trained model in a task-specific manner.
Second, compared to graph pre-training models, end-to-end GNN models can sometimes achieve comparable or even better performance. This implies that the discrepancy between the pre-training and downstream tasks in these pre-training approaches obstructs the knowledge transfer from the former to the latter. In such a case, even with sophisticated pre-training, they cannot effectively promote the performance of downstream tasks.
Third, the graph prompt model GPPT is only comparable to or even worse than the other baselines, despite also using prompts. A potential reason is that GPPT requires much more learnable parameters in their prompts than ours, which may not work well given very few shots (\eg, 1-shot). 

\stitle{Few-shot graph classification.}
We further conduct few-shot graph classification on four datasets, \ie, \emph{PROTEINS}, \emph{COX2}, \emph{ENZYMES}, and \emph{BZR}. 
For each dataset, we randomly generate 100 5-shot classification tasks for training and validation, following a similar process for node classification tasks. 

We illustrate the results of few-shot graph classification in Table~\ref{table.graph-classification}, and have the following observations.
First, our proposed \model\ significantly outperforms the baselines on these four datasets. This again demonstrates the necessity of unification for pre-training and downstream tasks, and the effectiveness of prompt-assisted task-specific aggregation for $\textsc{ReadOut}$.
Second, as both node and graph classification tasks share the same pre-trained model on \emph{PROTEINS} and \emph{ENZYMES}, the superior performance of \model\ on both types of task further demonstrates that, the gap between different tasks 
is well addressed by virtue of our unification framework.
Third, the graph pre-training models generally achieve better performance than the end-to-end GNN models.
This is because both InfoGraph and GraphCL capitalize on graph-level tasks for pre-training, which are naturally closer to the downstream graph classification. 

\begin{table}[tbp] 
    \centering
    \small
     \addtolength{\tabcolsep}{1mm}
    \caption{Accuracy evaluation on node classification.
    }
    \label{table.node-classification}%
    \vspace{-3mm} 
    {\footnotesize All tabular results are in percent, with best \textbf{bolded} and runner-up \underline{underlined}.}
    \\[1mm] 
    \begin{tabular}{@{}l|c|c|c@{}}
    \toprule
  \multirow{2}*{Methods} & \multicolumn{1}{c|}{Flickr} & \multicolumn{1}{c|}{PROTEINS} & \multicolumn{1}{c}{ENZYMES}  \\ & 50-shot & 1-shot & 1-shot  \\\midrule\midrule
    \method{GCN} &  9.22 $\pm$ 9.49 & 59.60 $\pm$ 12.44 & 61.49 $\pm$ 12.87 \\
    \method{GraphSAGE}  & 13.52 $\pm$ 11.28 & 59.12 $\pm$ 12.14 & 61.81 $\pm$ 13.19\\
     \method{GAT} & 16.02 $\pm$ 12.72 & 58.14 $\pm$ 12.05 & 60.77 $\pm$ 13.21\\
    \method{GIN} & 10.18 $\pm$ 5.41 & \underline{60.53} $\pm$ 12.19  & \underline{63.81} $\pm$ 11.28 
\\\midrule
    \method{DGI} & 17.71 $\pm$ 1.09 & 54.92 $\pm$ 18.46 & 63.33 $\pm$ 18.13\\
    \method{GraphCL}  & 18.37 $\pm$ 1.72 & 52.00 $\pm$ 15.83 & 58.73 $\pm$ 16.47\\
    \midrule
    \method{GPPT} & \underline{18.95} $\pm$ 1.92 & 50.83 $\pm$ 16.56 & 53.79 $\pm$ 17.46\\
    \midrule
    \method{GraphPrompt} & \textbf{20.21} $\pm$ 11.52 & \textbf{63.03} $\pm$ 12.14 & \textbf{67.04} $\pm$ 11.48\\
    \bottomrule
        \end{tabular}
\end{table}

\begin{table}[tbp] 
    \centering
    \small
     \addtolength{\tabcolsep}{-1mm}
    \caption{Accuracy evaluation on graph classification.}
    \vspace{-2mm}
    \label{table.graph-classification}%
    \begin{tabular}{@{}l|c|c|c|c@{}}
    \toprule
  \multirow{2}*{Methods} & \multicolumn{1}{c|}{PROTEINS} & \multicolumn{1}{c|}{COX2} & \multicolumn{1}{c|}{ENZYMES} & \multicolumn{1}{c}{BZR}  \\ & 5-shot  & 5-shot & 5-shot & 5-shot \\\midrule\midrule
    \method{GCN} & 54.87 $\pm$ 11.20 & 51.37 $\pm$ 11.06  & 20.37 $\pm$ 5.24	 & 56.16 $\pm$ 11.07	\\
    \method{GraphSAGE} & 52.99 $\pm$ 10.57 & 52.87 $\pm$ 11.46	 & 18.31 $\pm$ 6.22 & 57.23 $\pm$ 10.95\\
     \method{GAT} & 48.78 $\pm$ 18.46 & 51.20 $\pm$ 27.93 & 15.90 $\pm$ 4.13 & 53.19 $\pm$ 20.61\\
    \method{GIN} & \underline{58.17} $\pm$ 8.58 & 51.89 $\pm$ 8.71	& 20.34 $\pm$ 5.01 & 57.45 $\pm$ 10.54\\
    \midrule
    \method{InfoGraph} & 54.12 $\pm$ 8.20 & 54.04 $\pm$ 9.45 & 20.90 $\pm$ 3.32 & 57.57 $\pm$ 9.93\\
    \method{GraphCL} & 56.38 $\pm$ 7.24  & \underline{55.40} $\pm$ 12.04  & \underline{28.11} $\pm$ 4.00  & \underline{59.22} $\pm$ 7.42	\\
    \midrule
    \method{GraphPrompt} & \textbf{64.42} $\pm$ 4.37  & \textbf{59.21} $\pm$ 6.82 & \textbf{31.45} $\pm$ 4.32	 & \textbf{61.63} $\pm$ 7.68 \\\bottomrule
    \end{tabular}
\end{table}

\begin{figure}[t]
\centering
\includegraphics[width=1.0\linewidth]{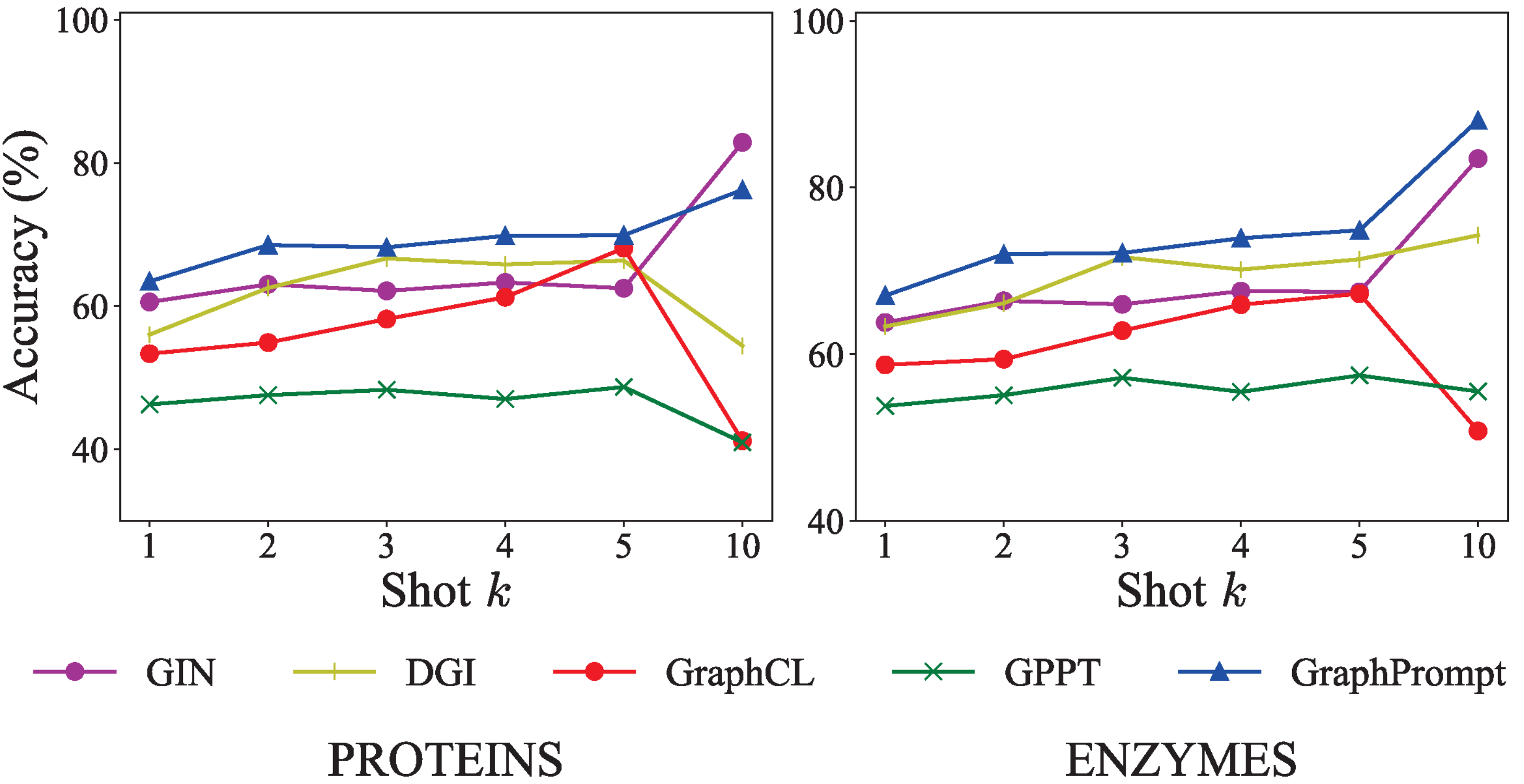}
\vspace{-3mm}
\caption{Impact of shots on few-shot node classification.}
\label{fig.node-few-shot-tune}
\end{figure}

\begin{figure}[t]
\centering
\includegraphics[width=1.0\linewidth]{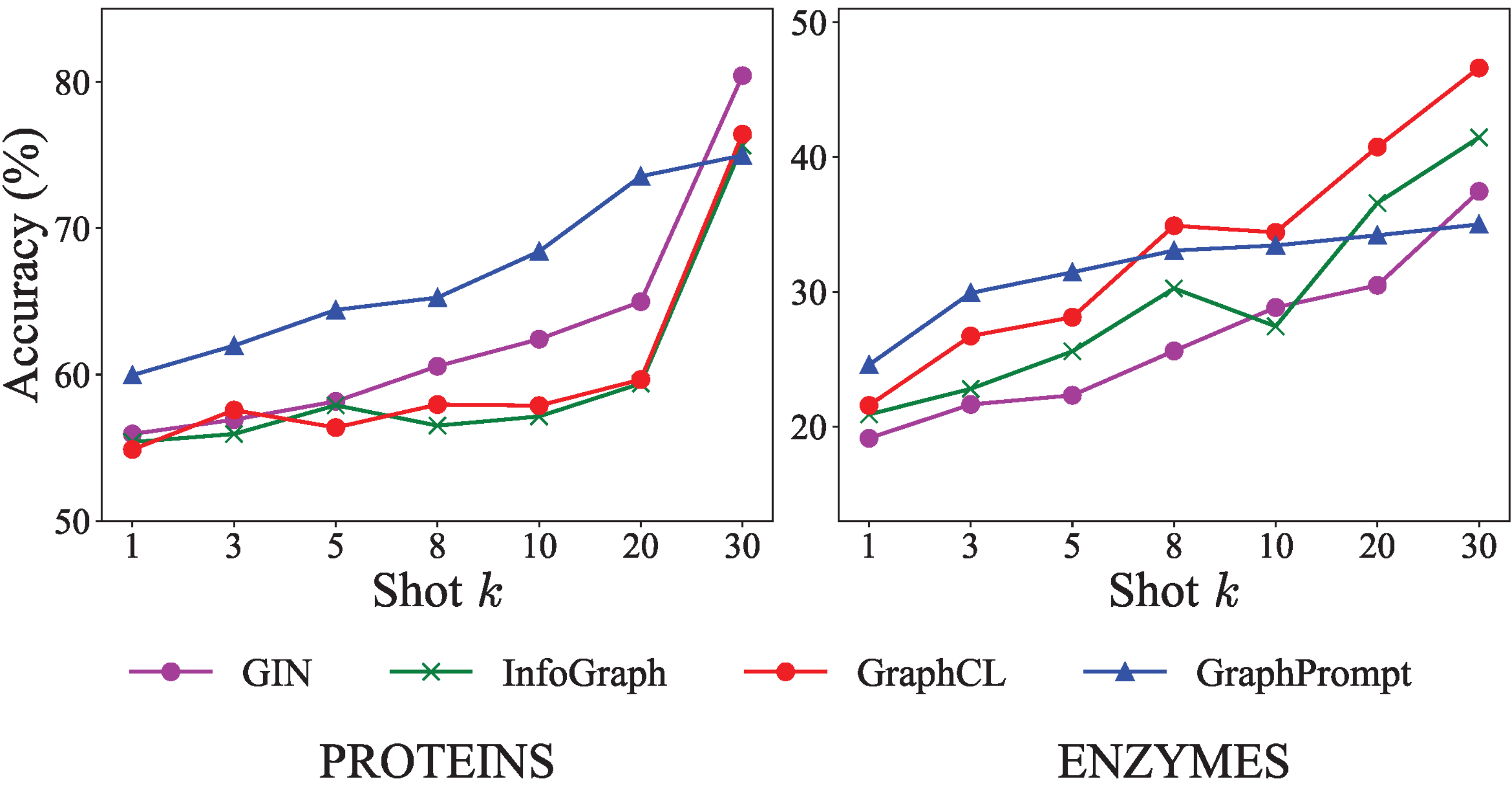}
\vspace{-3mm}
\caption{Impact of shots on few-shot graph classification.}
\label{fig.graph-few-shot-tune}
\end{figure}

\stitle{Performance with different shots.}
We study the impact of number of shots on the \emph{PROTEINS} and \emph{ENZYMES} datasets. For node classification, we vary the number of shots between 1 and 10, and compare with several competitive baselines (\ie, GIN, DGI, GraphCL, and GPPT) in Fig.~\ref{fig.node-few-shot-tune}.
For few-shot graph classification, we vary the number of shots between 1 and 30, and compare with competitive baselines (\ie, GIN, InfoGraph, and GraphCL) in Fig.~\ref{fig.graph-few-shot-tune}.
The task settings are identical to those stated earlier. 

In general, our proposed \model\ consistently outperforms the baselines especially with lower shots. For node classification, as the number of nodes in each graph is relatively small, 10 shots per class might be sufficient for semi-supervised node classification. Nevertheless, \model\ is competitive even with 10 shots. For graph classification, \model\ can be surpassed by some baselines when given more shots (\eg, 20 or more), especially on \emph{ENZYMES}.
On this dataset, 30 shots per class implies 30\% of the 600 graphs are used for training, which is not our target scenario. 



\subsection{Model Analysis}
We further analyse several aspects of our model. Due to space constraint, we only report the ablation and parameter efficiency study, and leave the rest to Appendix~\ref{app.experiments}.

\begin{figure}[t]
  \centering
  \subfigure[Node classification]{\label{fig.node-ablation}\includegraphics[width=0.48\linewidth]{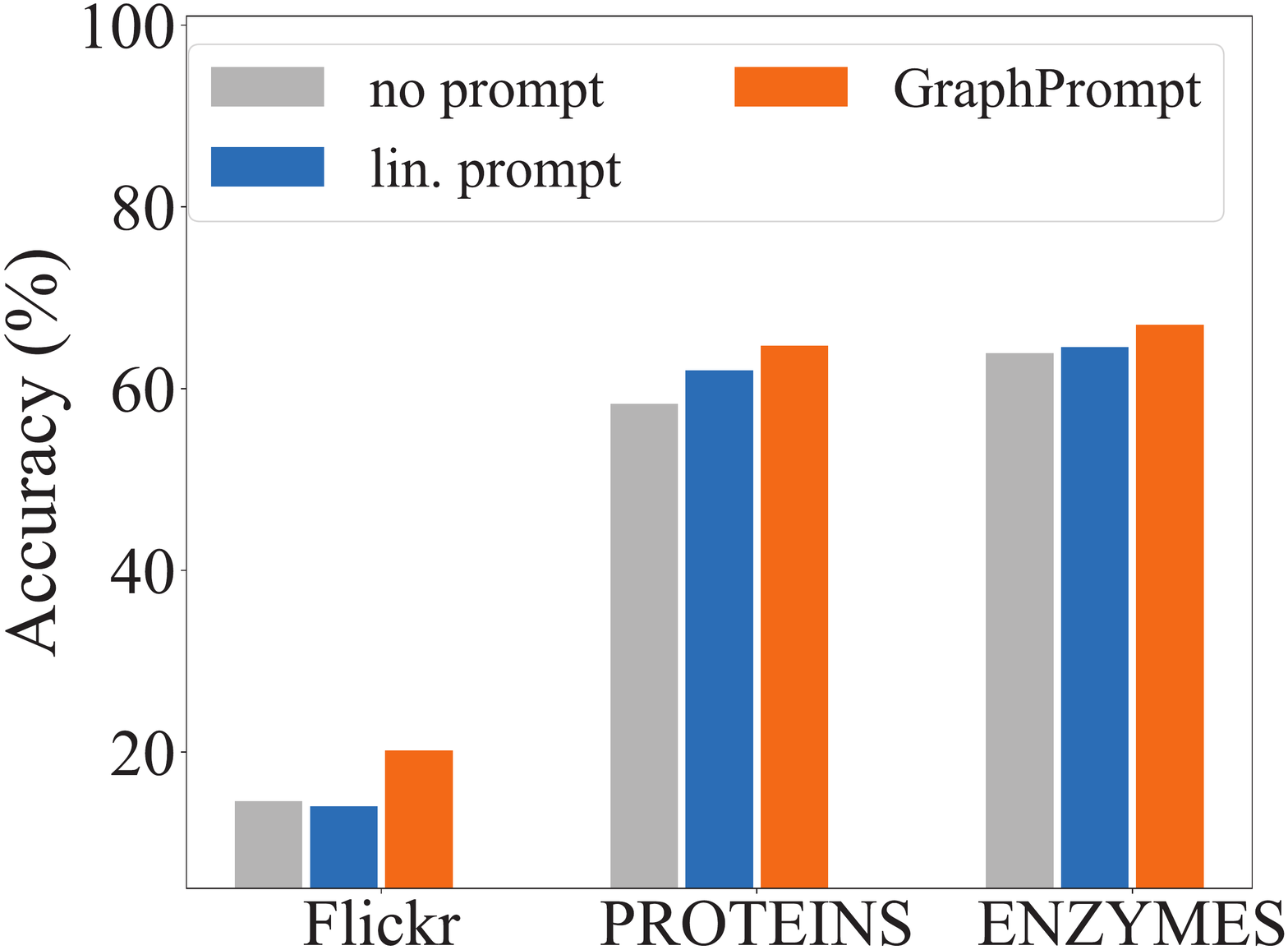}}
  \subfigure[Graph classification]{\label{fig.graph-ablation}\includegraphics[width=0.48\linewidth]{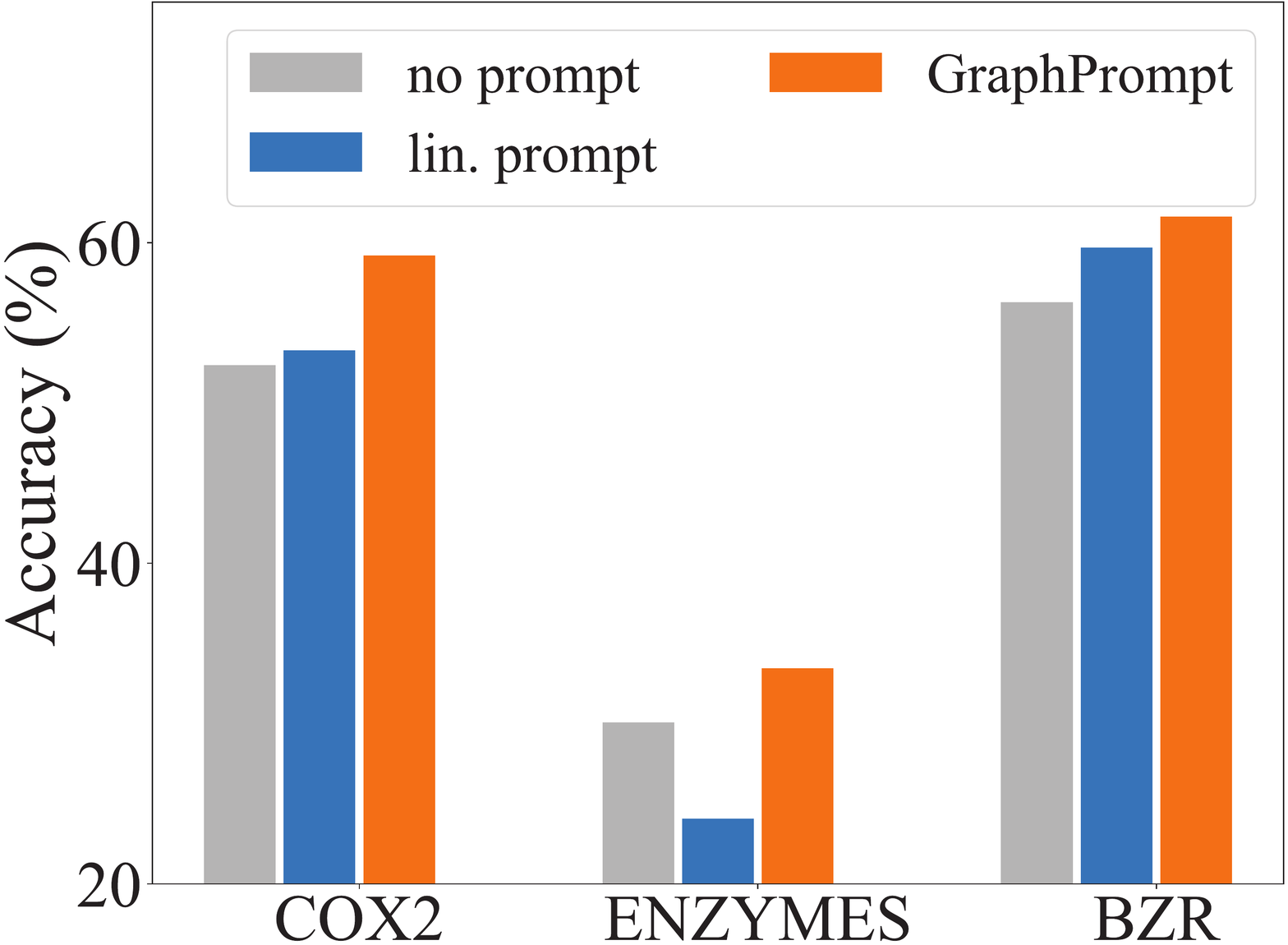}}
     \vspace{-3mm}
  \caption{Ablation study.} 
\label{fig.ablation}
\end{figure}

\stitle{Ablation study.}
To evaluate the contribution of each component, we conduct an ablation study by comparing \model\ with different prompting strategies:
(1) \emph{no prompt}: for downstream tasks, we remove the prompt vector, and conduct classification by employing a classifier on the subgraph representations obtained by a direct sum-based $\textsc{ReadOut}$.
(2) \emph{lin. prompt}: we replace the prompt vector with a linear transformation matrix in Eq.~\eqref{eq:prompt-lin}.

We conduct the ablation study on three datasets for node classification (\emph{Flickr}, \emph{PROTEINS}, and \emph{ENZYMES}) and graph classification (\emph{COX2}, \emph{ENZYMES}, and \emph{BZR}), respectively, and illustrate the comparison in Fig.~\ref{fig.ablation}. We have the following observations.
(1) Without the prompt vector, \emph{no prompt} usually performs the worst among the variants, showing the necessity of prompting the $\textsc{ReadOut}$ operation differently for different downstream tasks.
(2) Converting the prompt vector into a linear transformation matrix also hurts the performance, as the matrix involves more parameters thus increasing the reliance on labeled data.

\stitle{Parameter efficiency.}
We also compare the number of parameters that needs to be updated in a downstream node classification task  for a few representative models, as well as their number of floating point operations (FLOPs), in Table~\ref{table.parameters-num}.

In particular, as GIN works in an end-to-end manner, it is obvious that it involves the largest number of parameters for updating. 
For GPPT, it requires a separate learnable vector for each class as its representation, and an attention module to weigh the neighbors for aggregation in the structure token generation. Therefore, GPPT needs to update more parameters than \model, which is one factor that impairs its performance in downstream tasks.
For our proposed \model, it not only outperforms the baselines GIN and GPPT as we have seen earlier, but also requires the least parameters and FLOPs for downstream tasks. For illustration, in addition to prompt tuning,
if we also fine-tune the pre-trained weights instead of freezing them (denoted \model+ft), there will be significantly more parameters to update.



\begin{table}[!t] 
    \centering
    \small
    \addtolength{\tabcolsep}{-1.5pt}
   \caption{Study of parameter efficiency on node classification.}
    \begin{tabular}{@{}l|cc|cc|cc@{}}
    \toprule
  \multirow{2}*{Methods} & \multicolumn{2}{c|}{Flickr} & \multicolumn{2}{c|}{PROTEINS} & \multicolumn{2}{c}{ENZYMES}\\
  & Params & FLOPs & Params & FLOPs & Params & FLOPs \\\midrule\midrule
    GIN & 22,183 &240,100 & 5,730 & 12,380 & 6,280 & 11,030\\
    GPPT & 4,096 & 4,582 & 1,536 & 1,659 & 1,536 & 1,659 \\ \midrule
    \model\ & 96 &96 & 96 &96 & 96 &96   \\
    \model-ft &21,600 &235,200 &6,176 &13,440 &6,176 &10,944\\
\bottomrule
     \end{tabular}
      \label{table.parameters-num}%
\end{table}

\section{Conclusions}

In this paper, we studied the research problem of prompting on graphs and proposed \model, in order to overcome the limitations of graph neural networks in the supervised or ``pre-train, fine-tune'' paradigms.
In particular, to narrow the gap between pre-training and downstream objectives on graphs, we introduced a unification framework by mapping different tasks to a common task template. Moreover, to distinguish task individuality and achieve  task-specific optima, we proposed a learnable task-specific prompt vector that guides each downstream task to make full of the pre-trained model.  
Finally, we conduct extensive experiments on five public datasets, and show that \model\ significantly outperforms various state-of-the-art baselines.

\section*{Acknowledgments}
This research / project is supported by the Ministry of Education, Singapore, under its Academic Research Fund Tier 2 (MOE-T2EP20122-0041). Any opinions, findings and conclusions or recommendations expressed in this material are those of the author(s) and do not reflect the views of the Ministry of Education, Singapore.

\newpage

\balance


\bibliographystyle{ACM-Reference-Format}
\bibliography{references}

\clearpage
\appendix
\section*{Appendices}
\renewcommand\thesubsection{\Alph{subsection}}
\renewcommand\thesubsubsection{\thesubsection.\arabic{subsection}}
\setcounter{secnumdepth}{5}
\renewcommand{\thesection}{\Alph{section}}


\subsection{Algorithm and Complexity Analysis} \label{app.alg}
\stitle{Algorithm.}
We present the algorithm for prompt design and tuning of \model\ in Alg.~\ref{alg.prompt}. In line 1, we initialize the prompt vector and the objective $\bL_{\text{prompt}}$. 
In lines 2-3, we obtain the node embeddings of input graphs based on the pre-trained GNN. In lines 5-13, we accumulate the loss for the given tuning samples. In particular, in lines 5-6, we design the prompt for the specific task $t$. In lines 7-8, we calculate the subgraph representation for each class prototype. Then, in lines 9-13, we calculate and accumulate the loss and get the overall objective. Finally, in line 14 we optimize the prompt vector by minimizing the objective $\bL_{\text{prompt}}$.


\begin{algorithm}[h]
\small
\caption{\textsc{Prompt Design and Tuning}}
\label{alg.prompt}
\begin{algorithmic}[1]
    \Require Graphs set $\bG=\{G_j|j=1,2,\ldots\}$, task $t$-specific subgraphs set $\bS=\{S_{t,x}|x=1,2,\ldots\}$, 
    labeled set $\bD=\{(x_i,y_i)|i=1,2,\ldots\}$, class set $Y$,
    pre-trained GNN model $f_{\Theta_0}$ which takes in a graph and outputs its node embedding vectors.
    \Ensure Prompt vector $\vec{p}_t$. 
    \State $\vec{p}_t\leftarrow$ prompt vector initialization, $\bL_{\text{prompt}}\leftarrow 0$;
    \For{each graph $G_j\in \bG$}\Comment{Load pre-trained GNN}
        \State $\vec{H}_j \leftarrow f_{\Theta_0}(G_j)$
    \EndFor
    \While{not converged} \Comment{Tuning iteration}
        \For{each subgraph $s_t,_x\in \bS$}\Comment{Prompt design, Eq.~(12)}
            \State $\vec{s}_t,_x \leftarrow \textsc{ReadOut}(\{\vec{p}_t\odot\vec{h}_v:v\in V(S_x)\})$
        \EndFor
        \For {each class $c \in Y$}\Comment{Class prototypical subgraph}
            \State $\tilde{\vec{s}}_{t,c} \leftarrow $ Mean
            of node/graph embedding vectors
        \EndFor
        \For{each labeled pair $(x_i,y_i) \in \bD$}\Comment{Accumulate loss, Eq.~(14)}
            \State $\vec{Z}_i \leftarrow 0$
            \For{each class $c \in Y$}
                \State $\vec{Z}_i=\exp(\text{sim}(\vec{s}_{t,x_i},\tilde{\vec{s}}_{t,c})/\tau)+\vec{Z}_i$
            \EndFor
            \State $\bL_{\text{prompt}}=\bL_{\text{prompt}}-\ln({\exp(\text{sim}(\vec{s}_{t,x_i},\tilde{\vec{s}}_{t,y_i})/\tau)}/{\vec{Z}_i})$
        \EndFor
        \State Update $\vec{p}_t$ by minimize $\bL_{\text{prompt}}$;
    \EndWhile
    \State \Return $\vec{p}_t$.
\end{algorithmic}
\end{algorithm}

\stitle{Complexity analysis.}
For a node $v$, with average degree $\bar{d}$, $k$ GNN layers, $\delta$ hops for subgraph extraction, $D$ hidden dimensions, the complexity of GNN-based embedding calculation is $O(D\cdot \bar{d}^k)$, and the complexity of subgraph extraction is $O(\bar{d}^\delta)$. Thus, the embedding calculation of $v$'s subgraph with \textsc{ReadOut} is 
$O(D\cdot \bar{d}^k \cdot \bar{d}^\delta)$, where $k,\delta$ are small constants.
Furthermore, if some neighborhood sampling \cite{hamilton2017inductive} is adopted during GNN aggregation, 
$\bar{d}$ is a relatively small constant too.

\subsection{Further Descriptions of Datasets} \label{app.dataset}

We provide further details of the datasets.

(1) \emph{Flickr} \cite{wen2021meta} is an image sharing network,
which is collected by SNAP\footnote{\url{https://snap.stanford.edu/data/}}. 
In particular, each node is an image, and there exists an edge between two images if they share some common properties, such as commented by the same user, or from the same location. Each image belongs to one of the 7 categories.

(2) \emph{PROTEINS} \cite{borgwardt2005protein} is a collection of protein graphs which include the amino acid sequence, conformation, structure, and features such as active sites of the proteins. The nodes represent the secondary structures, and each edge depicts the neighboring relation in the amino-acid sequence or in 3D space. The nodes belong to three categories, and the graphs belong to two classes.

(3) \emph{COX2} \cite{nr} is a dataset of molecular structures including 467 cyclooxygenase-2 inhibitors, in which each node is an atom, and each edge represents the chemical bond between atoms, such as single, double, triple or aromatic. All the molecules belong to two categories.

(4) \emph{ENZYMES} \cite{wang2022faith} is a dataset of 600 enzymes collected from the BRENDA enzyme database. These enzymes are labeled into 6 categories according to their top-level EC enzyme.

(5) \emph{BZR} \cite{nr} is a collection of 405 ligands for benzodiazepine receptor, in which each ligand is represented by a graph. All these ligands belong to 2 categories.

Note that we conduct node classification on \emph{Flickr}, \emph{PROTEINS} and \emph{ENZYMES}, since their node labels generally appear on all the graphs, which is suitable for the setting of few-shot node classification on each graph. Note that, we only choose the graphs which consist of more than 50 nodes for the downstream node classification, to ensure there exist sufficient labeled nodes for testing. 
Additionally, graph classification is conducted on \emph{PROTEINS}, \emph{COX2}, \emph{ENXYMES} and \emph{BZR}. We use the given node features in the cited datasets to initialize input feature vectors, without additional processing.

\subsection{Further Descriptions of Baselines}

In this section, we present more details for the baselines, which are chosen from three main categories.

\vspace{1mm}
\noindent (1) \emph{End-to-end graph neural networks.}
\begin{itemize}[leftmargin=*]
    \item GCN \cite{kipf2016semi}: GCN resorts to mean-pooling based neighborhood aggregation to receive messages from the neighboring nodes for node representation learning in an end-to-end manner.
    \item GraphSage \cite{hamilton2017inductive}: GraphSAGE has a similar neighborhood aggregation mechanism with GCN, while it focuses more on the information from the node itself.
    \item GAT \cite{velivckovic2017graph} : GAT also depends on neighborhood aggregation for node representation learning in an end-to-end manner, while it can assign different weights to neighbors to reweigh their contributions.
    \item GIN \cite{xu2018powerful}: GIN employs a sum-based aggregator to replace the mean-pooling method in GCN, which is more powerful in expressing the graph structures. 
\end{itemize}

\noindent (2) \emph{Graph pre-training models.}
\begin{itemize}[leftmargin=*]
    \item DGI \cite{velickovic2019deep}: DGI capitalizes on a self-supervised method for pre-training, which is based on the concept of mutual information (MI). It maximizes the  MI between the local augmented instances and the global representation.
    \item InfoGraph \cite{sun2019infograph}: InfoGraph learns a graph-level representation, 
    which maximizes the MI  between the graph-level representation and substructure representations at various scales.
    \item GraphCL \cite{you2020graph}: GraphCL applies different graph augmentations to exploit the structural information on the graphs, and aims to maximize the agreement between different augmentations for graph pre-training.
\end{itemize}

\noindent (3) \emph{Graph prompt models.}
\begin{itemize}[leftmargin=*]
    \item GPPT \cite{sun2022gppt}. GPPT pre-trains a GNN model based on the link prediction task, and employs a learnable prompt to reformulate the downstream node classification task into the same format as link prediction.
\end{itemize}

\subsection{Further Implementation Details} \label{app.setting-parameter}

For baseline GCN \cite{kipf2016semi}, we employ a 3-layer architecture, and set the hidden dimension  as 32.
For GraphSAGE \cite{hamilton2017inductive}, we utilize the mean aggregator, and employ a 3-layer architecture. The hidden dimension is also set to 32.
For GAT \cite{velivckovic2017graph}, we employ a 2-layer architecture and set the hidden dimension as 32. Besides, we apply 4 attention heads in the first GAT layer.
Similarly, for GIN \cite{xu2018powerful}, we also employ a 3-layer architecture and set the hidden dimension as 32.
For the pre-training and prompting approaches, we use the backbones in their original paper.
Specifically, for DGI \cite{velickovic2019deep}, we use a 1-layer GCN as the backbone, and set the hidden dimension as 512. Besides, we utilize PReLU as the activation function.
For InfoGraph \cite{sun2019infograph}, we use a 3-layer GIN as the backbone, and set its hidden dimension as 32. 
For GraphCL \cite{you2020graph}, we also employ a 3-layer GIN as its backbone, and set the hidden dimension as 32. In particular, we choose the augmentations of node dropping and subgraph, with a default augmentation ratio of 0.2.
For GPPT \cite{sun2022gppt}, we utilize a 2-layer GraphSAGE as its backbone, set its hidden dimension as 128, and utilize the mean aggregator.
For our proposed \model, we employ a 3-layer GIN as the backbone, and set the hidden dimensions as 32. In addition, we set $\delta=1$ to construct 1-hop subgraphs for the nodes.


\subsection{Further Experimental Results} \label{app.experiments}

\begin{figure}[tbh]
\centering
\includegraphics[width=0.55\linewidth]{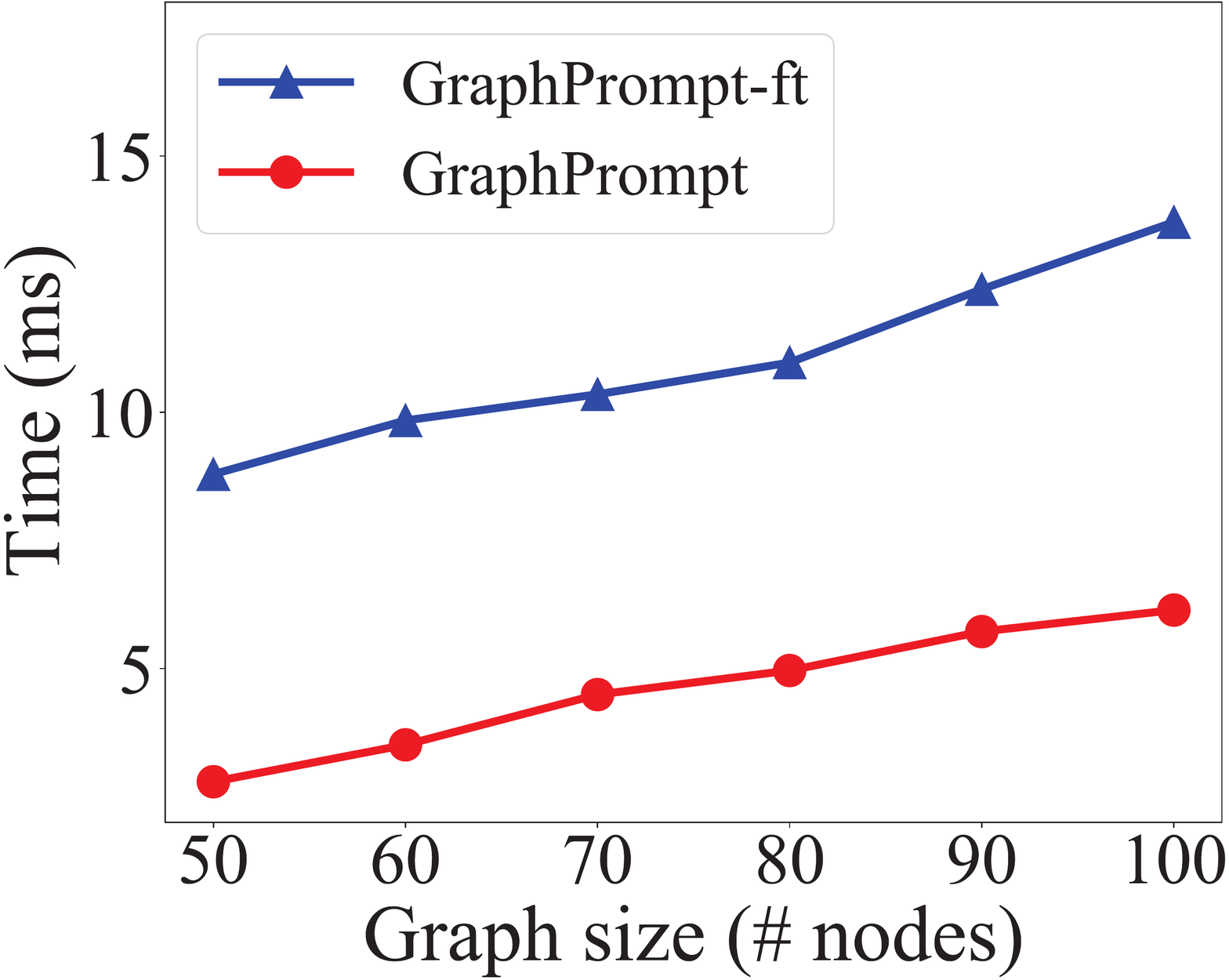}
\vspace{-2mm}
\caption{Scalability study.}
\label{fig.scalability}
\end{figure}

\stitle{Scalability study.}
We investigate the scalability of \model\ on the dataset \emph{PROTEINS} for graph classification. We divide the graphs into six groups based on their size (\ie, number of nodes). The size of graphs in each group is approximately 50, 60, $\ldots$, 100 nodes. We sample 10 graphs from each group, and record the prompt tuning time on the 10 graphs in each epoch. The results are presented in Fig.~\ref{fig.scalability}. Note that we also report the tuning time for \model-ft, a variant of \model, which fine-tunes all the parameters including the pre-trained GNN weights.
We first observe that the tuning time of our \model\ increases linearly as the graph size increases, demonstrating the scalability of \model\ on larger graphs. In addition, compared to \model, \model-ft needs more tuning time, showing the inefficiency of the fine-tuning paradigm.

\begin{figure}[htb]
  \centering
  \subfigure[Number of hops]{\label{fig.node-params-tune-K}\includegraphics[width=0.48\linewidth]{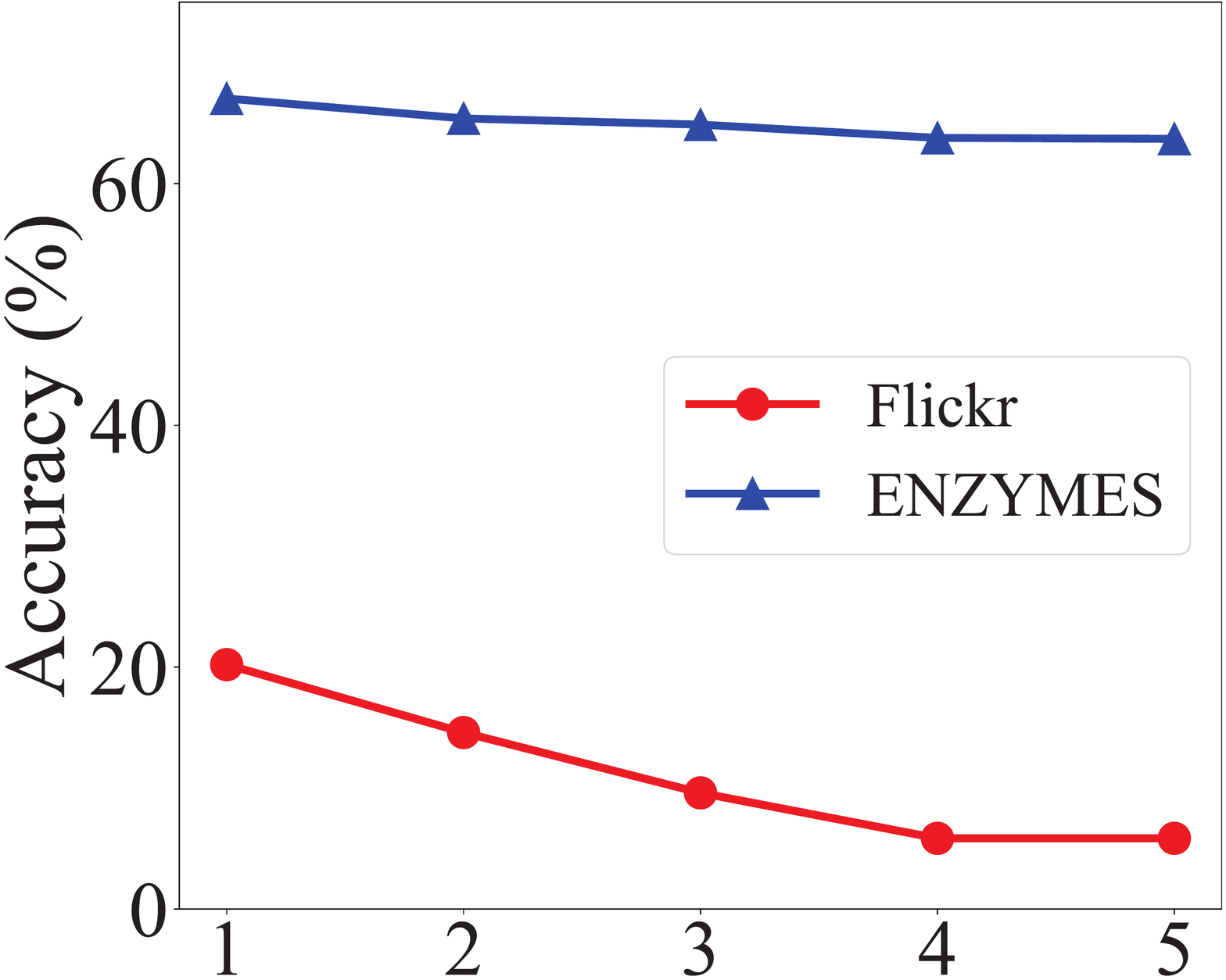}}
  \subfigure[Hidden dimension]{\label{fig.node-params-tune-d}\includegraphics[width=0.48\linewidth]{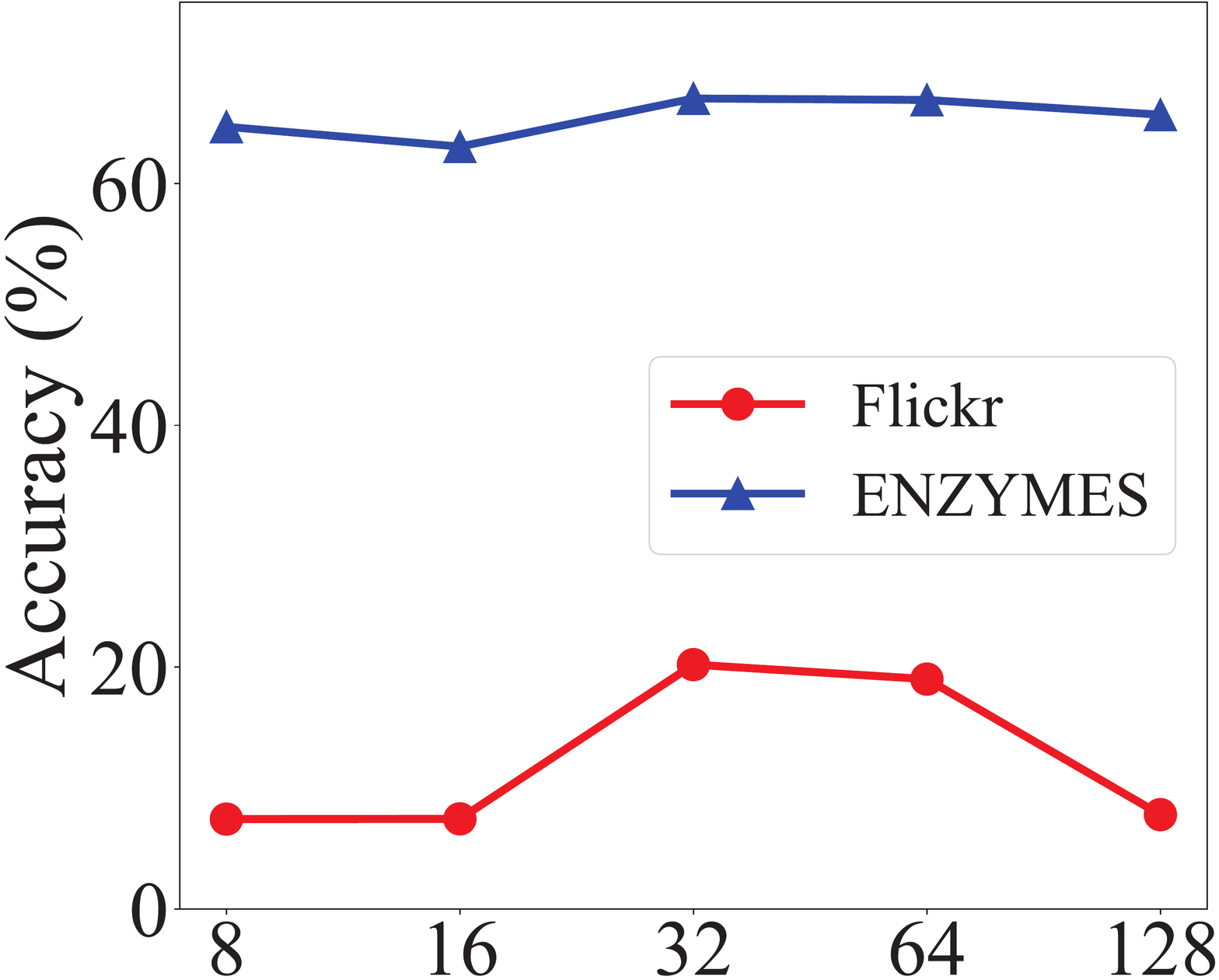}}
     \vspace{-2mm}
  \caption{Parameter sensitivity on node classification.} 
\label{fig.tune-node}
\end{figure}

\begin{figure}[htb]
  \centering
  \subfigure[Number of hops]{\label{fig.graph-params-tune-K}\includegraphics[width=0.48\linewidth]{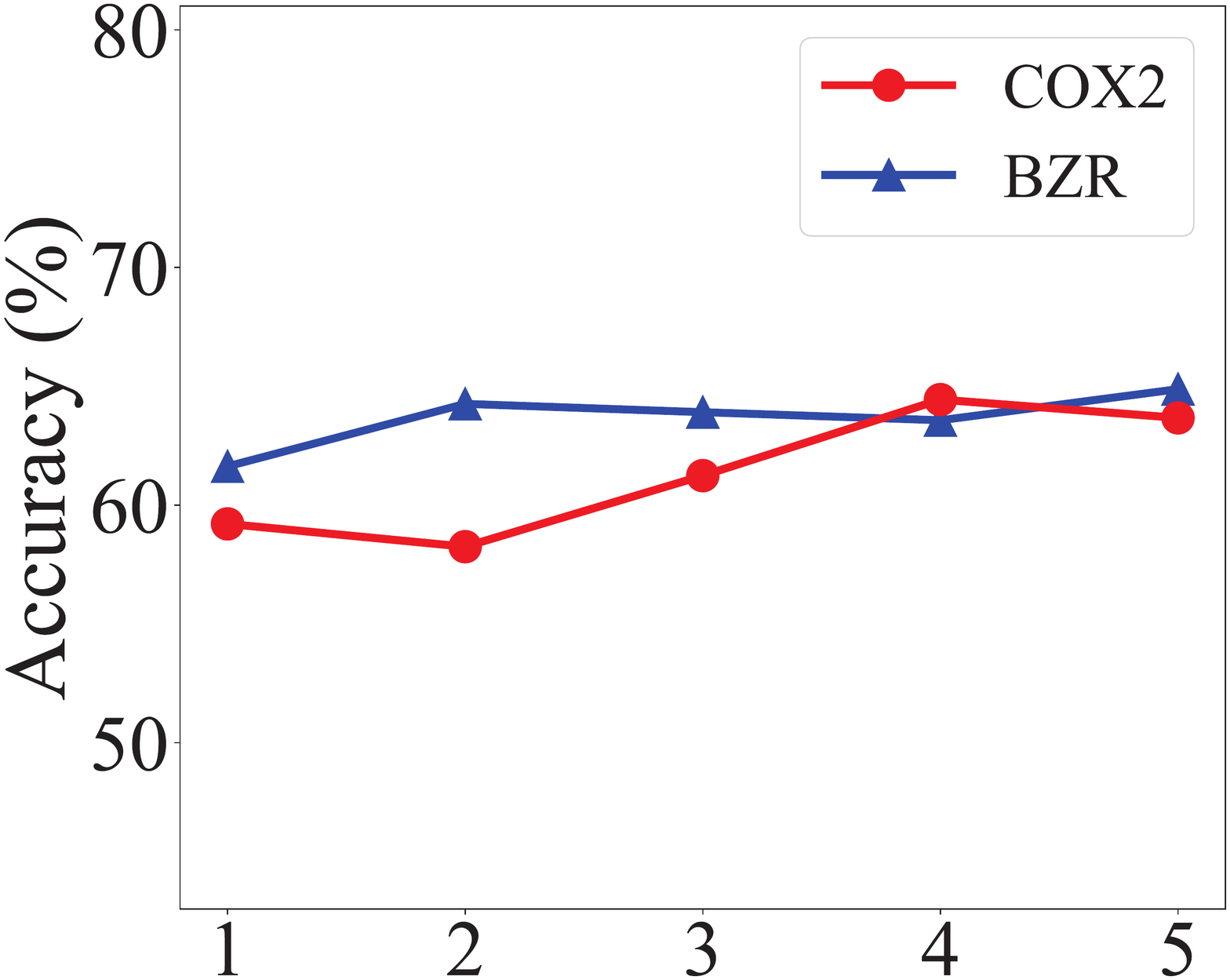}}
  \subfigure[Hidden dimension]{\label{fig.graph-params-tune-d}\includegraphics[width=0.48\linewidth]{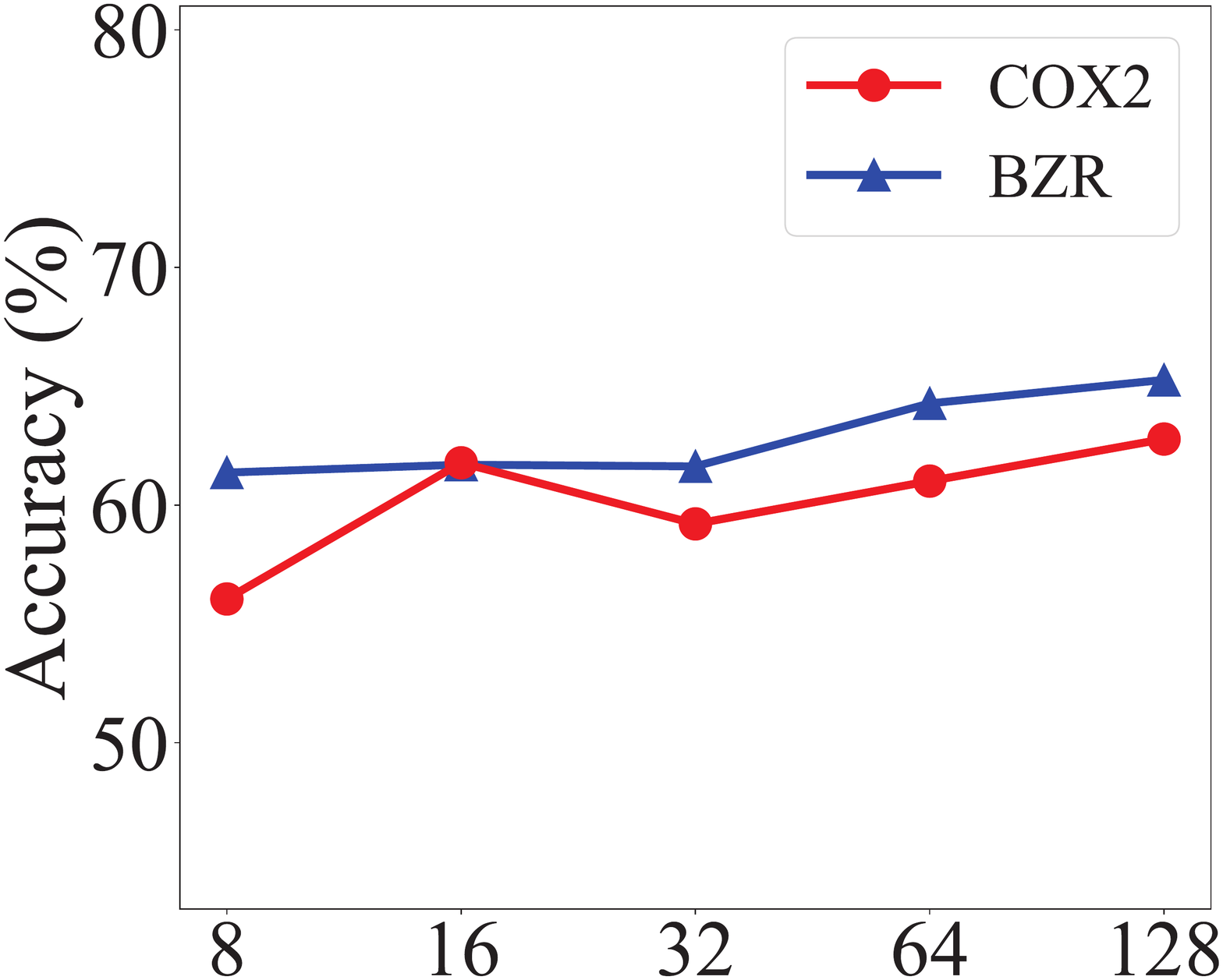}}
     \vspace{-2mm}
  \caption{Parameter sensitivity on graph classification.} 
\label{fig.tune-graph}
\end{figure}

\stitle{Parameter sensitivity.}
We evaluate the sensitivity of two important hyperparameters in \model, and show the impact in Figs.~\ref{fig.tune-node} and \ref{fig.tune-graph} for node classification and graph classification, respectively.

For the number of hops ($\delta$) in subgraph construction, the performance on node classification gradually decreases as the number of hops increases. This is because a larger subgraph tends to bring in irrelevant information for the target node, and may suffer from the over-smoothing issue \cite{chen2020measuring}. On the other hand, for graph classification, the number of hops only affects the pre-training stage as the whole graph is used in downstream classification. In this case, the number of hops does not show a clear trend, implying less impact on graph classification since both small and large subgraphs are helpful in capturing  substructure information at different scales.

For the hidden dimension, a smaller dimension is better for node classification, such as 32 and 64. For graph classification, a slightly larger dimension might be better, such as 64 and 128. Overall, 32 or 64 appears to be robust  for both node and graph classification.

\subsection{Data Ethics Statement}
To evaluate the efficacy of this
work, we conducted experiments which only use publicly available
datasets, namely, Flickr\footnote{\url{https://snap.stanford.edu/data/web-flickr.html}}, PROTEINS, COX2, ENZYMES and BZR\footnote{\url{https://chrsmrrs.github.io/datasets/}}, in accordance to their usage terms
and conditions if any.
We further declare that no personally identifiable information was
used, and no human or animal subject was involved in this research.

\end{document}